\documentclass[letterpaper, 10pt, conference]{ieeeconf}     
\IEEEoverridecommandlockouts                              
\usepackage{cite}
\usepackage{mathtools} 
\usepackage{amsmath} 
\usepackage{amssymb}  
\usepackage{siunitx}
\usepackage{textcomp}
\usepackage{xcolor}
\usepackage[hidelinks]{hyperref}
\usepackage{cleveref}
\usepackage{subcaption}
\usepackage{graphicx}
\usepackage{float}
\graphicspath{{Images/}{Images/draft/}} 
\usepackage{booktabs}
\usepackage{tabularx}
\usepackage{tikz}
\usetikzlibrary{positioning, fit, backgrounds, arrows.meta}
\definecolor{ETH-l-blue}{RGB}{205,214,230} 
\usepackage{framed}
\usepackage{soul}
\usepackage{xcolor}
\usepackage[normalem]{ulem} 
\usepackage{dblfloatfix} 
\usepackage{color}

\usepackage[font=small,skip=3pt]{caption}

\begin{document}

\title{\LARGE \bf All You Need Is Low Fidelity: Zero-Shot Sim-to-Real of Learned Robotic Fish Control}

\author{Liam Maloney$^{*}$, Simon Ramchandani$^{*}$, Mike Y. Michelis, Ronan Hinchet, Robert K. Katzschmann%
\thanks{$^{*}$Equal contribution.}%
\thanks{All authors are with the Soft Robotics Lab, ETH Zurich, Switzerland. M.~Y.~Michelis is also with the ETH AI Center, ETH Zurich, Switzerland.}%
}

\maketitle

\thispagestyle{empty}
\pagestyle{empty}

\begin{abstract}
Complex tasks for underwater robots remain limited by the capabilities of their controllers. Learning a better one for a soft, underactuated robotic fish trades simulator cost against fidelity. We show that an intentionally low-fidelity simulator is enough: a stateless, quasi-steady fluid model with no wake and no added-mass history suffices to learn a \emph{general}, closed-loop controller that transfers to hardware without tuning. Our platform is a soft, single-motor, tendon-driven fish whose policy observes only what the hardware can measure. A staged pipeline grounds the simulator in two independent identifications, fixing the tail dynamics and a stateless fluid model; the policy then acts through a band-limited rhythmic trajectory generator rather than commanding the tail directly. Deployed unchanged in an outdoor pool, a single policy performs closed-loop target reaching, disturbance rejection, and out-of-distribution target acquisition and tracking. The transfer rests on the constraint rather than the fidelity: the generator cannot leave the band over which the fluid was identified. This raises the question of how much of the physics can reside in the controller rather than in the simulator.
\end{abstract}



\section{Introduction}
\label{sec:introduction}

Soft robotic fish promise quiet, agile platforms for observing marine life~\cite{katzschmann_exploration_2018}. While rigid quadrupeds and humanoids have learned whole-body controllers that produce a wide range of versatile motions~\cite{kumar2021}, soft swimming robots have largely cruised in a straight line or executed a set of hard-coded swimming gaits. We ask whether a soft, underactuated robotic swimmer can also be given general learned whole-body control deployed on real hardware.

\begin{figure}[t]
      \centering
      \includegraphics[width=\columnwidth]{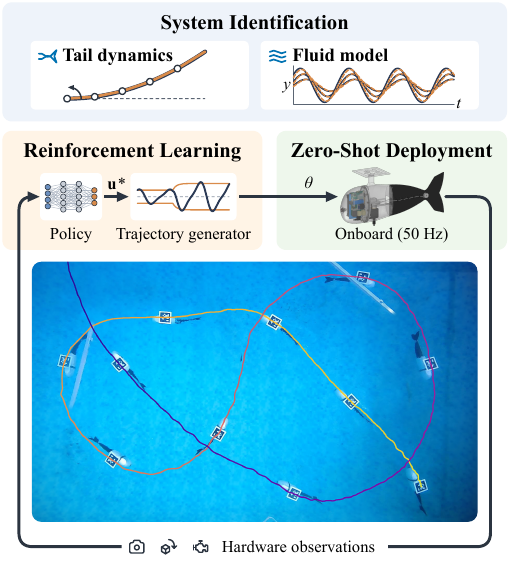}
       \caption{Overview. Separate experiments identify the tail dynamics and a stateless fluid model. The policy sets the parameters $\mathbf{u}^*$ of a band-limited trajectory generator (TG) that emits the motor setpoint $\theta_m$. The same policy and TG run unchanged onboard the real soft fish robot, closed by hardware observations.}
      \label{fig:overview}
      \vspace{-2pt}
  \end{figure}

Reaching that generality is hard precisely because of the compliance that makes these robots attractive. Our platform, based on Obayashi et al.~\cite{obayashi2025scafi}, is \emph{underactuated}, its tail shape emerging from the coupled dynamics of the motor, elastic body, tendon routing, and a nonlinear fluid rather than being commanded joint by joint. Feedback control on top of hand-designed gaits can steer the stroke but not change it, and here the stroke itself must adapt to the task. Reinforcement learning (RL) offers a route towards robust control policies, but requires millions of interactions with no comparable real-world dataset, so training must happen in simulation.

A computational fluid dynamics (CFD) solver simulates swimming physics faithfully, and agile policies have been trained inside CFD~\cite{lin_learning_2025}, but resolving the flow is costly per step and hard to parallelize, a poor fit for the large sample budgets of RL. The fast alternative is a simplified simulator such as MuJoCo~\cite{todorov_mujoco_2012}, whose \emph{stateless}, \emph{quasi-steady} hydrodynamics make learning feasible, but only coarsely approximate a real unsteady fluid. The open question is whether a controller learned in such a low-cost simulator can be transferred to a physical fish at all, and made general rather than tuned to a single gait.

\begin{figure*}[!t]
      \centering
      \includegraphics[width=\textwidth]{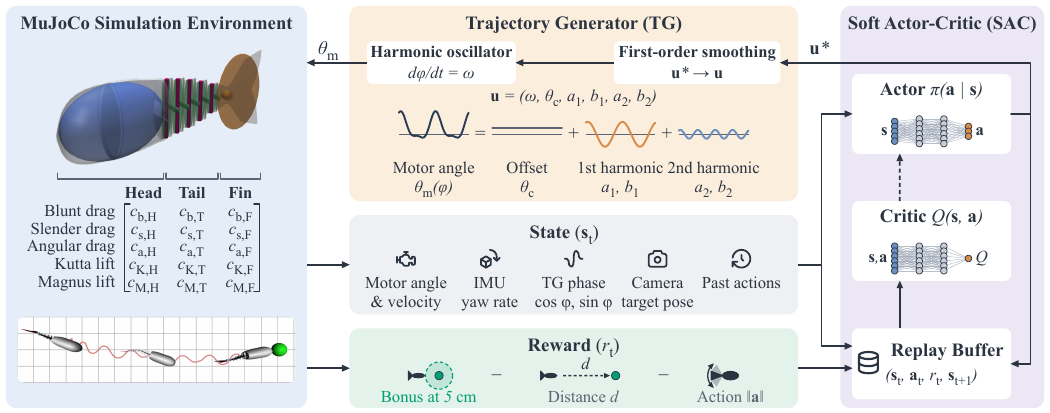}
      \caption{Policy training loop. \textbf{Left:} the MuJoCo twin and the fifteen quasi-steady fluid coefficients identified for its three zones (\Cref{sec:sysid-fluid}, \Cref{tab:fluid}). \textbf{Center:} the policy commands target gait parameters $\mathbf{u}^*$ rather than motor angles. These are smoothed by a first-order time lag $\tau$ into $\mathbf{u}$, which the trajectory generator converts into motor setpoint $\theta_m$ (\Cref{eq:tj}). The observed state holds only what the hardware can measure. \textbf{Right:} soft actor-critic, rewarded for closing the distance to the target (\Cref{eq:reward}).}
      \label{fig:pipeline}
  \end{figure*}

We show that this low-fidelity simulator is enough, provided the policy is held inside the regime its fluid coefficients were identified on (\Cref{fig:overview}). Rather than command the tail directly, our policy sets the parameters of a trajectory generator (TG), whose frequency ceiling is that regime, and is trained through reinforcement learning (\Cref{fig:pipeline}) entirely in the fast simulator. A single such policy transfers to the physical robot zero-shot and closes the loop on real tasks: target reaching, disturbance rejection, and out-of-distribution target acquisition and tracking. To our knowledge this is the first closed-loop swimming policy trained in a \emph{stateless} fluid model and deployed on a soft underactuated fish. The exchange should outlast the platform: a simulator need only be accurate over the motions its controller can produce, which puts learned control on compliant robots within reach of anyone without a validated flow solver.
\section{Related Work}
\label{sec:related-work}

Training such a controller requires a simulator, and the field spans a wide fidelity-cost range: CFD that resolves the unsteady wake~\cite{borazjani_numerical_2008}, GPU-accelerated fluid-structure platforms built for training~\cite{liu_fishgym_2022}, and fast, stateless quasi-steady models that approximate hydrodynamic force from the instantaneous velocity of a body~\cite{vaxenburg2025whole, michelis2026}. Whatever the fidelity, a sim-to-real gap remains, narrowed by domain randomization~\cite{tobin2017} and by system identification on a differentiable simulator~\cite{zhang_sim2real_2022}.

We likewise train under domain randomization, but the choice of fidelity also shapes what the controller can become: trained in an accurate CFD solver, an end-to-end policy transfers zero-shot to agile maneuvers~\cite{lin_learning_2025}, at the cost of hours of simulation per policy and a multi-actuated body with per-joint feedback. By contrast, in a low-cost stateless simulator Wang et al.~\cite{wang2022} deploy a policy on an underactuated wire-driven eel, but only for straight-line swimming. Where learning has been applied to target tracking on a real fish, the authority of the policy is often limited to a single steering offset added to a hand-tuned gait~\cite{yu2021}. The digital twin of Michelis et al.~\cite{michelis2026} uses a stateless fluid model as well and trains a general target-reaching policy on a tendon-driven fish, but in simulation only and with privileged tail-state access that the compliant hardware cannot measure. We build toward the combination that these leave open: a policy with authority over the whole gait (\Cref{sec:control-tg}), learning from body and motor state alone (\Cref{sec:control-rl}) in a stateless fluid simulator (\Cref{sec:setup-sim}), deployed for general closed-loop control on a soft underactuated fish (\Cref{sec:transfer}).

\section{Platforms}
\label{sec:setup}

\subsection{Hardware Fish}
\label{sec:setup-hw}

\begin{figure}[!b]
    \centering
    \includegraphics[width=\columnwidth]{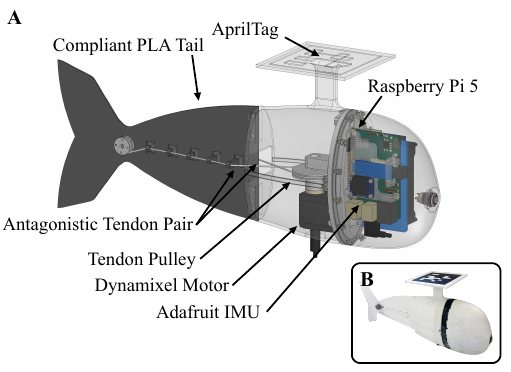}
    \caption{\textbf{A:} Translucent CAD model of the hardware fish. A sealed, oil-filled head carries the onboard compute; the motor in the flooded body turns the pulley that drives the antagonistic tendon pair along the compliant tail. \textbf{B:} The assembled fish as deployed, with the AprilTag on the head that the overhead camera tracks.}
    \label{fig:real-sofi}
\end{figure}

Our platform is a tendon-driven soft robotic fish of the body-and-caudal-fin type (\Cref{fig:real-sofi}). A Raspberry Pi~5 runs the policy from within a sealed 3D-printed head filled with mineral oil to reduce overall buoyancy. In the central flooded body of the fish, a Dynamixel XW430-T200-R motor operated in position control rotates the drive pulley. A pair of antagonistic tendons is wrapped in opposing directions on the pulley and routed along the tail: a compliant \SI{1.8}{\milli\meter} 3D-printed PLA plate. The pulley provides control of both stroke amplitude and frequency, bending the tail into a traveling S-shape. The tail is \emph{underactuated}: one actuator drives the continuum. A motor encoder, Adafruit BNO055 IMU, and 3D-printed AprilTag provide the policy observations.


\subsection{Simulation Fish}
\label{sec:setup-sim}

Policies are trained in MuJoCo~\cite{todorov_mujoco_2012}, using a digital fish twin~\cite{michelis2026}, with its geometry redesigned to match our hardware (\Cref{fig:pipeline}). This model is fast enough to carry the sample budget RL needs, the reason we do not instead resolve the flow with CFD. The fish model consists of a rigid head plus a five-hinge tail chain, where meshes exported from the CAD assembly give the model its appearance, but the hydrodynamics act only on an approximated ellipsoid geometry. The tapered PLA tail is divided into rigid sections joined by five rotational hinges with torsional spring-dampers. The antagonistic tendon pair is modeled as a \emph{spatial} tendon through that same routing, so the one measured shaft stiffness sets all five joint stiffnesses rather than leaving them free (\Cref{sec:sysid-tail}); damping is assigned proportionally. Segment masses, centers of mass, and moments of inertia are read directly off the CAD assembly rather than identified.

Fluid forces use the ellipsoid-based stateless model of MuJoCo, a 3D generalization of the quasi-steady force decomposition derived for falling cards~\cite{andersen_analysis_2005, vaxenburg2025whole}. Each ellipsoid carries five coefficients, and the fluid force depends only on the \emph{instantaneous} velocity of the body, so there is no fluid state and therefore no wake history. We extend the fluid from five coefficients to fifteen: three independent sets identified in \Cref{sec:sysid-fluid}. The motor is a position-control servo at the shaft (\Cref{sec:sysid-motor}).


\section{System Identification}
\label{sec:sysid}


\subsection{Tail Dynamics}
\label{sec:sysid-tail}

\begin{figure}[!t]
    \centering
    \includegraphics[width=\columnwidth]{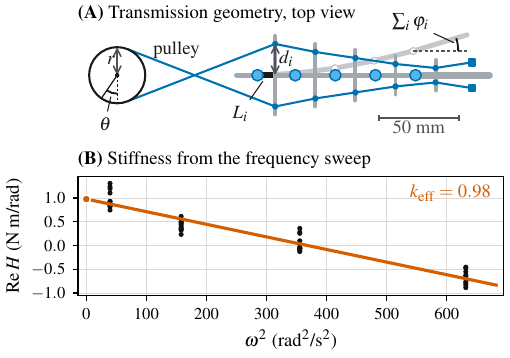}
    \caption{Tail dynamics. \textbf{A:} Transmission geometry in top view, to scale from the deployed model. The antagonistic tendon pair runs from the pulley of radius $r$ through a guide bracket at each of the five hinges, $i=1$ at the root to $5$ at the fin, at moment arm $d_i$, and terminates at the squares. $L_i$ is the flexure between a hinge axis and its bracket. Grey is the same tail under load, at the static equilibrium of the identified twin; $\theta$ is drawn enlarged and the $\varphi_i$ accumulate to $\sum_i\varphi_i$. \textbf{B:} $\operatorname{Re}H$ against $\omega^2$, $42$ drive conditions over the four frequencies in air; the intercept is $k_\mathrm{eff}$.}
    \label{fig:tailstiff}
\end{figure}

The first identification fixes the passive dynamics of the tail, its stiffness and damping. Driven through the motor shaft (\Cref{fig:tailstiff}A), the tail is a damped oscillator, $\tau = k_\mathrm{eff}\theta + c_\mathrm{eff}\dot\theta + I_\mathrm{eff}\ddot\theta$. Writing $\theta(t)=\operatorname{Re}[\Theta\,\mathrm{e}^{\mathrm{j}\omega t}]$ gives the torque-to-angle frequency response
\begin{equation}
    H(\omega) \;=\; \frac{T}{\Theta} \;=\; \underbrace{\big(k_\mathrm{eff}-I_\mathrm{eff}\,\omega^2\big)}
    _{\operatorname{Re}H}
    \;+\; \mathrm{j}\,\underbrace{c_\mathrm{eff}\,\omega}_{\operatorname{Im}H} ,
    \label{eq:tailstiff}
\end{equation}
so $\operatorname{Re}H$ is affine in $\omega^2$ and a straight line through the drive frequencies gives the structural stiffness as its intercept and the effective inertia as its slope. We sweep the tail with sinusoids at four frequencies in air, free of any fluid load. Reading stiffness from a frequency response is established practice~\cite{celic2008}, and on this actuator it is also the only clean one: friction likewise acts along velocity, so it falls into $\operatorname{Im}H$ and cannot contaminate the stiffness, unlike a static load-deflection test, which traces a wide hysteresis loop instead. $\operatorname{Im}H$ then gives the damping from the same sweep, once its amplitude-dependent Coulomb part is separated from the viscous one, which gives $c_\mathrm{eff}\approx\SI{0.011}{\newton\meter\second\per\radian}$. The intercept is $k_\mathrm{eff}\approx\SI{0.98}{\newton\meter\per\radian}$ (\Cref{fig:tailstiff}B), against $0.26$ for the idealized model felt at the same shaft.

That is a single number at the motor, and the model needs one stiffness per hinge. The tendon relates the two (\Cref{fig:tailstiff}A): under tendon tension $F$ joint $i$ deflects $\varphi_i = F d_i/k_i$, the arcs $\varphi_i d_i$ sum to the length the pulley pays out, and $\tau = F r$, so
\begin{equation}
    k_\mathrm{eff} \;=\; \frac{r^2}{\sum_{i} d_i^{2}/k_i} ,
    \label{eq:keffmap}
\end{equation}
with $r=\SI{17.5}{\milli\meter}$ and moment arms from $19.5$ to \SI{4.5}{\milli\meter}. The five hinges share modulus and thickness but taper in height. Euler--Bernoulli bending leaves $k_i \propto h_i/L_i$: a shape fixed by geometry alone, carrying one scale that the measured $k_\mathrm{eff}$ sets. The resulting per-joint stiffness runs $2.99$ to \SI{1.32}{\newton\meter\per\radian} from body to fin, with damping assigned proportionally. Fixing the stiffness first, and from in-air data alone, matters for what follows: an under-stiff tail leaves a restoring torque that a fluid fit can only absorb by inflating its angular-drag coefficients.

\subsection{Fluid Model}
\label{sec:sysid-fluid}

Where the predecessor shared one set of five coefficients across the fish and suggested per-segment tuning as future work~\cite{michelis2026}, we split the fluid into three zones, head and body, tail, and caudal fin, giving the fifteen coefficients of \Cref{tab:fluid}. We fit them from open-loop free-swim runs with the motor decoupled: recorded motor positions are replayed, so the coefficients never depend on the motor model. With one shared set, the high angular drag that stabilizes the yaw oscillation of the body would also lock the caudal fin, which must oscillate freely to make thrust. Giving each zone its own coefficients resolves this. The identification is a trajectory-fitting problem solved with CMA-ES~\cite{hansen2016}. The fin angular-drag and Kutta-lift coefficients come out an order of magnitude larger than those of the head (\Cref{tab:fluid}). Because a stateless model cannot represent the unsteady vortex-driven thrust of a real fin, the fit inflates the coefficients. This inflation, which we return to in \Cref{sec:control}, is intrinsic to the model rather than a fitting error: a separate identification on different data inflates the fin just as much.

\begin{figure}[!ht]
    \centering
    \includegraphics[width=\columnwidth]{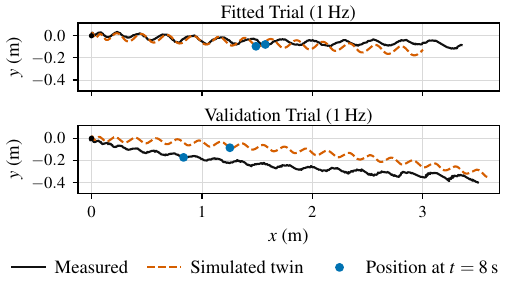}
    \caption{Fluid fit: measured against the identified twin, for one fitted and one validation run at the same condition and on the same axes. The circles mark the position at $t=\SI{8}{\second}$ on both paths.}
    \label{fig:fluidfit}
\end{figure}

The fit deliberately weights the acceleration phase: position errors at the start of a run carry four times the weight of those at the end. Left unweighted, the fitted twin swims backwards for the first strokes before momentum builds, which the real fish never does. While a body accelerates it must also push the surrounding fluid aside, an added-mass effect~\cite{wise2018acceleration}. In our model that term is fixed by the ellipsoid geometry rather than fitted~\cite{vaxenburg2025whole}, so what remains of the transient is absorbed into the fitted coefficients by the same mechanism that inflates the fin. Held-out runs test whether the weighting identified the fluid or merely the transient (\Cref{fig:fluidfit}).

\begin{table}[!ht]
    \centering
    \caption{Identified quasi-steady fluid coefficients, five per zone. The angular drag and Kutta lift of the fin are markedly larger, the stateless model inflating them to stand in for unsteady thrust.}
    \label{tab:fluid}
    \small
    \setlength{\tabcolsep}{4pt}
    \begin{tabular*}{\columnwidth}{@{\extracolsep{\fill}} lccccc}
    \toprule
    Zone & blunt & slender & angular & Kutta & Magnus \\
    \midrule
    Head/Body & $0.006$ & $0.045$ & $0.813$  & $0.521$  & $0.903$ \\
    Tail & $0.039$ & $0.551$ & $5.428$  & $0.001$  & $1.581$ \\
    Fin  & $0.171$ & $0.004$ & $18.310$ & $10.148$ & $5.951$ \\
    \bottomrule
    \end{tabular*}
    \vspace{-8pt}
\end{table}

\subsection{Motor}
\label{sec:sysid-motor}
\begin{figure}[!h]
    \centering
    \includegraphics[width=\columnwidth]{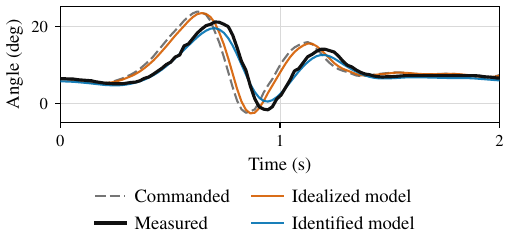}
    \caption{Motor shaft angle over a representative stroke: commanded setpoint, measured hardware, the idealized model the deployed policy was trained on, and the model identified afterwards.}
    \label{fig:motor-overlay}
    \vspace{-2pt}
\end{figure}

We identified the motor after the deployment, prompted by the question of whether an unconstrained policy would behave differently against a realistic actuator. The result also bears on the deployed policy: it shows how far the actuator it was trained on differs from the real one. In the position-controlled drive the motor torque balances the identified structural response of the tail against a position servo,
\begin{align}
    \tau_\mathrm{motor} &= k_\mathrm{eff}\,\theta + c_\mathrm{eff}\,\dot\theta + I_\mathrm{eff}\,\ddot\theta
    \nonumber \\
    &= k_p\big(\theta_\mathrm{ref}(t-\tau_d) - \theta\big) + \tau_\mathrm{fluid} 
    \label{eq:motor},
\end{align} 
where the rotor armature raises the effective inertia, the velocity-limiting damping raises $c_\mathrm{eff}$, and the servo contributes a finite gain $k_p$ and a transport delay $\tau_d$. The model we used during deployment treated this servo as effectively ideal: near-rigid in position, near-massless, and instantaneous; the physical Dynamixel is weak, heavy, speed-limited by back-EMF, and delayed. Those same limits rule out the frequency response used for the tail: saturation and a back-EMF speed cap make the response amplitude-dependent, so we identify in the time domain instead, free-running the model on the recorded command, fitting by CMA-ES against the measured encoder angle, and validating on held-out commands~\cite{schoukens2019}. The identified servo (\Cref{tab:motor}) reproduces the motion to $1.0^\circ$ RMSE against $3.6^\circ$ for the idealized model (\Cref{fig:motor-overlay}). The policy that swam in the pool had therefore been trained against an actuator bearing little resemblance to the one it commanded, which we take up in \Cref{sec:discussion}.

\begin{table}[!h]
    \centering
    \caption{Motor actuator parameters: the idealized model the deployed policy was trained on, against the motor model identified afterwards. Position gain and rotor armature are fitted; the force limit is the hardware value; the joint damping caps peak shaft speed near the measured \SI{5}{\radian\per\second}.}
    \label{tab:motor}
    \small
    \begin{tabular*}{\columnwidth}{@{\extracolsep{\fill}} lcc}
    \toprule
    Parameter & Idealized & Identified \\
    \midrule
    Position gain $k_p$              & $800$   & $5.41$ \\
    Velocity gain $k_v$              & $20$    & $0$ \\
    Rotor armature [\si{\kilo\gram\meter\squared}]       & $0.001$ & $0.0113$ \\
    Joint damping $b$ [\si{\newton\meter\second\per\radian}]  & $0$     & $0.48$ \\
    Force limit [\si{\newton\meter}]          & $1.0$   & $2.3$ \\
    Transport delay $\tau_d$ [\si{\milli\second}]    & $0$     & ${\approx}14$ \\
    \bottomrule
    \end{tabular*}
    \vspace{-8pt}
\end{table}

\section{Control}
\label{sec:control}


\subsection{The Control Problem}
\label{sec:control-problem}

The quasi-steady, stateless fluid model computes thrust from instantaneous velocity alone,

\begin{equation}
    v = A \cdot 2\pi f, \qquad F_\mathrm{thrust} \propto v^2 .
    \label{eq:v-thrust}
\end{equation}

Consequently, a given thrust can be produced by a large, slow stroke or by a small, fast flutter; the model does not distinguish them. An unconstrained policy commanding the motor angle directly, \emph{direct action} below, converges on the flutter the model rewards but was never validated on. Bounding the motor velocity does not remove this, since the policy simply lowers the amplitude and raises the frequency to reach the same speed; nor does a low-pass filter on the setpoint: attenuating the amplitude only pushes the policy higher in frequency, and the velocity reaching the fluid rises with it, which we measure in \Cref{sec:transfer-why}. A real fish, by contrast, produces thrust through unsteady, vortex-driven motion with a visible, large-amplitude stroke. We therefore \emph{expect} the small-amplitude, high-frequency gaits that the stateless model rewards not to transfer, an expectation we qualify in \Cref{sec:discussion}. This points less to a better plant than to a constraint on how the policy may act.

\subsection{Trajectory Generator}
\label{sec:control-tg}

We supply this action constraint with a trajectory generator (TG). Rather than command the tail directly, the policy sets the parameters of a compact periodic signal that emits a smooth, rhythmic setpoint with a hard ceiling on its frequency enforced by its construction. The flutter is then simply unreachable, while the policy keeps authority over the whole gait: amplitude, offset and turning. Because a capped frequency can reach a given speed only with a larger amplitude, the TG steers the policy toward the visible, large-amplitude strokes a real fish uses. Its output is a low-order harmonic in the phase $\phi$,


\begin{equation}
    \theta_m = \theta_c + \sum_{k=1}^{2} \big[\, a_k \cos(k\phi) + b_k \sin(k\phi) \,\big],
    \label{eq:tj}
\end{equation}

whose phase advances at the commanded frequency, $\dot{\phi} = \omega$. The six TG parameters $\mathbf{u}=(\omega, \theta_c, a_{1}, b_{1}, a_{2}, b_{2})$ are not set directly but relaxed toward a policy-set target $\mathbf{u}^*$ with time constant $\tau$,

\begin{equation}
    \dot{\mathbf{u}} = \frac{1}{\tau} \left( \mathbf{u}^* - \mathbf{u} \right).
\end{equation}

We found $\tau = \SI{0.1}{\second}$ to result in stable, agile motions, where the parameters adapt within a stroke but never jump between steps. Neither piece is new: letting a policy modulate the parameters of a trajectory generator rather than the joint commands is an established action-space prior~\cite{iscen2018pmtg, bellegarda2022cpgrl, ijspeert2008}. What is specific here is the purpose of the frequency ceiling: it is set to the band the fluid coefficients were fit on (\Cref{sec:sysid-fluid}), so the action space cannot leave the region where the stateless model was validated. 

\subsection{Reinforcement Learning}
\label{sec:control-rl}


The policy is trained by soft actor-critic~\cite{haarnoja_soft_2018}, a standard off-policy method, using the implementation in Stable-Baselines3~\cite{stable-baselines3}. Both actor and critic are three-layer MLPs with 286 ReLU units per layer (\num{175}k actor parameters) and operate on the same observations. Neither receives privileged simulator states as input; only the reward uses the exact simulator distance. The 24-dimensional observation is limited to what the hardware can measure: the motor angle as its $\cos,\sin$ and its velocity, the body yaw rate from the IMU, the TG phase as $(\cos,\sin)$, the last three frequency/offset commands $(\omega,\theta_c)$, and the three most recent overhead-camera frames with each including the target distance, its offset vector in the fish frame, and how stale the frame is. The action is the 6-dimensional TG parameter vector $\mathbf{u}=(\omega, \theta_c, a_{1}, b_{1}, a_{2}, b_{2})$. The reward favors proximity to the target while lightly penalizing large actions, with a terminal bonus for arrival, where $d$ is the distance to the target.

\begin{equation}
    r = -d \;-\; 0.1\,\lVert a\rVert \;+\; 1000\cdot\mathbf{1}[\,d < 0.05~\mathrm{m}\,] ,
    \label{eq:reward}
\end{equation}

 Training follows a ten-stage curriculum, promoted at $>\!95\%$ rolling target acquisition. The schedule adds one difficulty at a time, in three phases that each introduce a new kind of demand: first observation noise and delay, then per-episode physics randomization, then the target arc widened from a narrow forward cone through $\pm60^\circ$ and $\pm120^\circ$ to $\pm180^\circ$. The deployed policy completed the schedule through the $\pm120^\circ$ stage, which is why targets beyond that band are out-of-distribution in \Cref{sec:transfer}. Domain randomization~\cite{peng2018} perturbs head-body mass, inertia and center of mass ($\pm5\%$, \SI{\pm5}{\milli\meter}), tail stiffness and damping ($\pm15\%$), and the fluid coefficients ($\pm20\%$). Sensor noise (encoder, IMU, AprilTag, vision latency and frame drops) was measured on hardware and rounded up, so the policy never trains against less noise than it meets in the pool. Within each phase the learning rate decays stepwise ($\times0.3$ per stage) and is reset to its initial value at each phase boundary, a warm-restart schedule~\cite{loshchilov2017sgdr} that lets the optimizer re-adapt to the step change in the task without flushing the replay buffer.
\section{Sim-to-Real Transfer}
\label{sec:transfer}


\begin{figure*}[tbp]
    \centering
    \includegraphics[width=\textwidth]{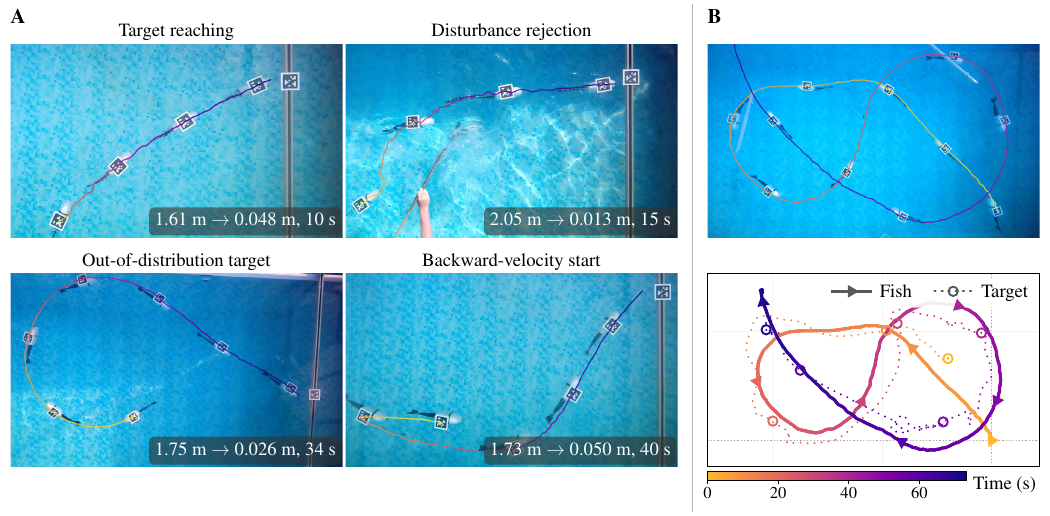}
    \caption{Compilation of paths swum by the deployed soft fish. \textbf{A:} Overhead pool composites of four fixed-target runs, each panel titled by the condition it tests, with fish snapshots at fixed intervals over the full swum path and the AprilTag-marked target. Notes read initial distance, closest approach and elapsed time. \textbf{B:} Dynamic tracking of a target moved by hand along a figure-eight for about \SI{73}{\second}. Top, the overhead composite; bottom, the tracked fish (solid) and target (dotted) paths, markers every \SI{10}{\second}.}
    \label{fig:deploy}
\end{figure*}

\subsection{Experimental Setup}
\label{sec:transfer-track}

During deployment, an overhead camera detects two AprilTags~\cite{olson2011}, one fixed to the fish and one acting as target, recovering the planar pose of both at roughly $60$~fps. The tag attached to the fish is also used to capture the resulting trajectory during each trial. A laptop connected to the camera processes and sends the three most recent poses to the Raspberry Pi~5. These, along with IMU and encoder states, are fed to the trained policy running at \SI{50}{\hertz}. This produces the requested TG parameters which the Pi~5 tracks to command the Dynamixel motor through position control.

\subsection{Deployment}
\label{sec:transfer-deploy}

\begin{figure}[!ht]
    \centering
    \includegraphics[width=\columnwidth]{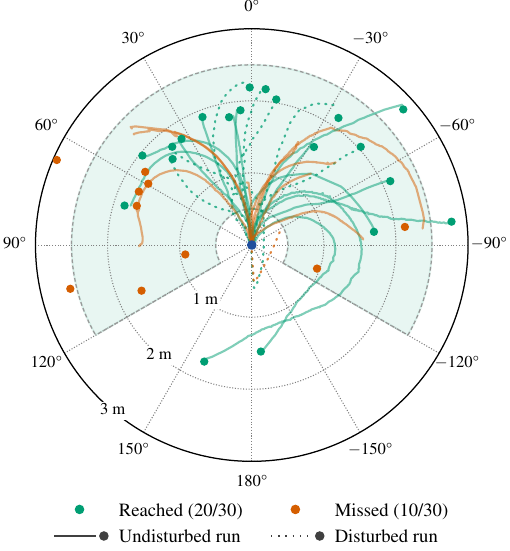}
    \caption{Reachability of the deployed policy with each target and swum path in the start frame of the fish (shaded wedge represents the trained region: $\pm120^\circ$, $0.5$--\SI{2.5}{\meter}) colored by outcome at the \SI{20}{\centi\meter} criterion. All 30 fixed-target runs are shown together (22 undisturbed and 8 disturbed). Misses concentrate in the lateral-rear $90$--$120^\circ$ band. Rear ($>\!\pm120^\circ$) and $>$\SI{2.5}{\meter} targets are reached.}
    \label{fig:reach-collage}
\end{figure}

The TG policy, trained only in the stateless simulator of \Cref{sec:setup,sec:sysid}, is deployed unchanged on the physical fish in an outdoor pool (${\sim}5.5\times2.7$~m) and closes the loop on targets (\Cref{fig:deploy}). From \SI{1.6}{\meter} the fish converges to within \SI{4.5}{\centi\meter} with a smooth, large-amplitude stroke. Across the session it arrived inside the \SI{5}{\centi\meter} training radius (\Cref{eq:reward}) on $11$ of $22$ static and $5$ of $8$ perturbed targets, and within \SI{20}{\centi\meter} on $13$ and $7$ (\Cref{fig:reach-collage}). Since targets were hand-placed, not sampled uniformly, these are descriptive session counts, not an unbiased reachability rate. On hardware the gait is clean and low-frequency, not the broadband flutter of direct action. The measured tail beat is \SI{1.3}{\hertz} across the deployment runs, well below the frequency ceilings.

\subsection{Beyond the Training Distribution}
\label{sec:transfer-ood}
The deployed policy was asked for several things it never saw in training, and did them. Its target distribution ended at $\pm120^\circ$ and \SI{2.5}{\meter}, yet it reached targets at $+158^\circ$ and $-175^\circ$, nearly directly behind, closing to $2.1$ and \SI{1.6}{\centi\meter} by a wide U-turn, and out to \SI{2.8}{\meter}, closing to \SI{2.3}{\centi\meter}. Trained only on static targets, it followed a hand-moved one in both attempts, a proof of concept for driving it with an automated path planner. Feedback also absorbs disturbances an open-loop gait could not: no external disturbance appeared in training, yet it reached the target in $7$ of $8$ perturbed trials, two of three released with an imposed backward or spun start, both prods with a stick, and all three swims through stirred water. Every trajectory generator scaling constant differed between training and the shipped configuration, amplitude and turn bias $2.7\times$ and $5.3\times$ smaller and the frequency ceiling $1.5\times$ higher, yet it still reached, recovered and tracked.

One limit recurred. The lateral-rear $90^\circ$--$120^\circ$ band went $0/4$, and only one of ten figure-eight attempts produced the two clean crossing loops shown. Both demand a tighter turn than the bias of the TG can close within a pool of this size; this is a turning-radius limit, not a tracking failure.

\subsection{Why the Transfer Works}
\label{sec:transfer-why}

A stateless fluid model is only as good as the motion it was identified on, so the question is not whether an unconstrained policy swims in simulation but whether we could tell if it did. Rolling out the deployed trajectory generator, unfiltered direct action, and direct action behind a first-order setpoint filter over the $\pm180^\circ$ task under domain randomization, we compare where each places its tail-velocity power against the $1$--\SI{4}{\hertz} band the fluid was identified over.

Velocity is the right diagnostic, not the commanded setpoint or the tail angle. A small, fast flutter barely moves the tail, so it looks harmless in the angle, but its velocity matches that of a large, slow stroke (\Cref{eq:v-thrust}), and velocity is what the fluid squares into thrust. The commanded setpoint is unusable here because it carries the \SI{50}{\hertz} stepping of the controller, which the motor removes and the fluid never sees.

\begin{figure}[!h]
    \centering
    \includegraphics[width=\columnwidth]{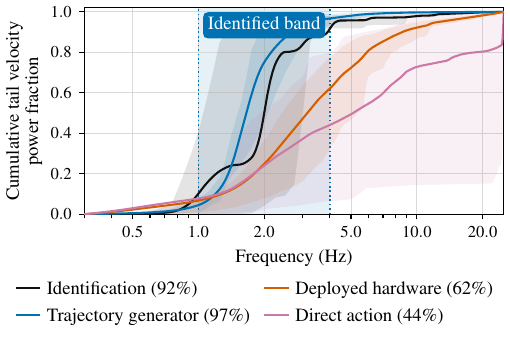}
    \caption{Cumulative tail-velocity power against frequency, from the measured motion. Each run is normalized to unit power before pooling per condition, so a run counts once regardless of length; shading is the $10$--$90\%$ spread across runs. The identification curve steps because its sweep is discrete. Legend percentages are the fraction below \SI{4}{\hertz}. Deployment spectra are resampled onto a uniform grid from the logged timestamps, as the identification pipeline already does. Twelve identification runs, $25$ simulated episodes per policy, $43$ deployment runs.}
    \label{fig:velspectrum}
\end{figure}

The open-loop identification runs place $92\%$ of their velocity power below \SI{4}{\hertz} (\Cref{fig:velspectrum}). The trajectory generator places $97\%$ there, so the policy operates in the regime the model was fit on. Unconstrained direct action places $44\%$. Its gait is not thereby wrong, but it is one we cannot assess: outside the identified band there is no measurement against which the fluid was checked, so we cannot say what its thrust prediction is worth and therefore cannot say whether such a gait would transfer.

On hardware the deployed policy places $62\%$ of its velocity power below \SI{4}{\hertz}, between the trajectory generator and unconstrained direct action; because the encoder is sampled asynchronously, residual broadband content remains and this is a lower bound. Part of that shortfall is a scaling error rather than a physical one: the frequency ceiling was set to \SI{6}{\hertz} at deployment against the \SI{4}{\hertz} used in training, and because the action maps linearly onto that ceiling every commanded frequency was $1.5\times$ higher than intended. Rescaling the measured spectra by that factor recovers $77\%$; the remainder we do not account for.

\begin{figure}[tbp]
    \centering
    \includegraphics[width=\columnwidth]{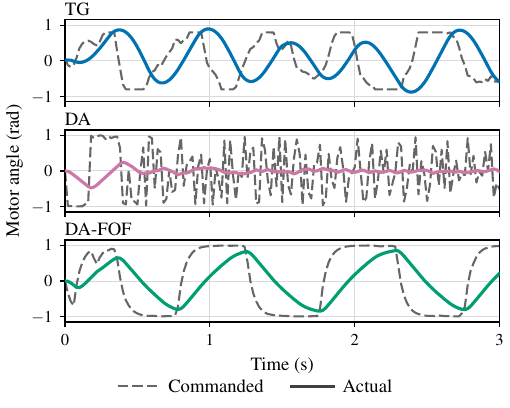}
    \caption{Commanded setpoint and measured motor angle for the trajectory generator (TG), unconstrained direct action (DA), and direct action with a first-order setpoint filter (DA-FOF). Direct action commands a rapid oscillation the motor does not reproduce.}
    \label{fig:gaits}
\end{figure}

A first-order filter on the setpoint, the obvious alternative, goes most of the way: over $40$ episodes it holds $93\%$ of velocity power in band while the fish swims forward, but only $68\%$ in the $14$ episodes where the gait reverses (\Cref{fig:gaits}). Attenuating a setpoint is not the same as bounding its frequency, which the trajectory generator does by construction, in both directions.

These are simulation results, and the three policies were trained against different fluid fits and different motor models, so the velocity-frequency signature is comparable across them but the comparison is not a controlled experiment.

\section{Discussion}
\label{sec:discussion}


One reading of \Cref{sec:control-problem} generalizes beyond this robot, though our experiments do not prove it: a reinforcement learning policy will exploit whatever its simulator cannot distinguish. A stateless fluid cannot separate a large slow stroke from a small fast flutter (\Cref{eq:v-thrust}), and an unconstrained policy converges on the flutter, the regime the coefficients were never fit to (\Cref{sec:sysid-fluid}). 

On this view a low-cost simulator is valid over a bounded region of state space, roughly the regime its coefficients were identified on, here moderate-amplitude, moderate-frequency undulation, and unreliable wherever the physics it omits, unsteady vortex shedding and added-mass transients chief among them, matters. An unconstrained policy has no notion of that boundary and every incentive to cross it, since anything the model cannot penalize is free reward. The band-limited TG keeps the policy inside that region by construction; we used it because it was the constraint available on a position-controlled pulley drive without reward engineering or an unsteady-flow solver. Whether tightening the fluid model instead would have transferred is something our experiments cannot settle: we never ran an unconstrained policy on the fish. A controlled deployment of both action spaces with a common plant would decide this.

A related fragility surfaced earlier in the pipeline, before any policy was involved. During early system identification, certain combinations of body mass distribution and fluid coefficients produced a simulated fish that swam net backward under a forward-driving gait: a sign error, not a magnitude one. Fixing the inertial properties from the CAD assembly (\Cref{sec:setup-sim}), rather than leaving them for the fluid fit to absorb, removed it. The lesson is the same at a different stage: a quasi-steady model fit without a firm prior on the parameters it is not meant to explain can compensate for structural error with implausible coefficients, up to the wrong sign of net thrust. Constraining what a parameter may absorb during identification is the analogue of constraining what a policy may command during control.

Why the policy reached beyond its training distribution follows from the same two choices. It never observes where the target is, only its offset in the fish frame, and is rewarded for closing that offset, so it learns a servo on an error rather than a map over the distribution it was trained on: further away, behind, or moving is a larger or time-varying error, not a new kind of input. The same structure is why so much could be wrong at once and not matter: the deployed policy was trained in a fluid with no wake, through an actuator model bearing little resemblance to the motor it would command, and shipped with every generator scaling constant wrong, and it still reached, recovered, and tracked. Because the generator bounds what it may emit, an unfamiliar error still produces a gait inside the family the fluid was identified on. What transfers to another problem is that split: keep the part that has to generalize small, a handful of gait parameters driven by an error, and the physics then only has to be correct over the motions those parameters can produce.

\section{Conclusion and Future Work}
\label{sec:conclusion}

A single RL policy, trained entirely in a cheap stateless fluid simulator, gives a real underactuated soft fish general closed-loop control, zero-shot: one trajectory-generator policy, deployed unchanged in an outdoor pool, reached static targets, rejected disturbances, generalized past its trained distribution, and tracked a moving one. The simulator is grounded in two independent identifications, the tail dynamics and a fifteen-coefficient fluid model, each fixed by its own experiment. The band-limited trajectory generator kept the learned gait inside the regime those identifications cover, where an unconstrained policy instead exploited the small-amplitude, high-frequency flutter a stateless fluid rewards. 

Three extensions follow. Identifying the validity boundary directly, and enforcing it with something less blunt than a fixed harmonic oscillator, may recover agility the present prior gives up. Replacing hand-placed targets with a sampled grid would turn our session counts into an unbiased reach map. Lastly, adding randomized external forces during training would build the disturbance rejection we observed but did not design for. The question this raises is not how much physics a simulator must capture, but how much of it can reside in the controller instead. If this trade holds beyond this fish, learned policies could reach soft robots long before faithful simulators arrive.




\bibliographystyle{IEEEtran}
\bibliography{bibliography}

\end{document}